\documentclass[letterpaper]{article}

\usepackage{aaai2027}

\usepackage[hyphens]{url}
\usepackage{graphicx}
\usepackage{natbib}
\usepackage{caption}
\usepackage{marvosym}
\usepackage{booktabs}
\usepackage{multirow}
\usepackage{amsmath,amssymb}
\usepackage{tikz}
\usetikzlibrary{arrows.meta,positioning,fit,shapes.geometric}

\newcommand{\methodname}{HiGram}

\title{Hierarchical Graph Memory for LLM Agents with  Path-level \\Localization and Rewrite}
\author{
    Xiawei Yue\textsuperscript{\rm 1, \rm 2},
    Boran Wang\textsuperscript{\rm 1, \rm 2},
    Xiaoqing Zhang\textsuperscript{\rm 2},
    Shuxin Zheng\textsuperscript{\rm 2},
    Ziwei Zhang\textsuperscript{\rm 3}\thanks{Corresponding author.}
}

\affiliations{
    \textsuperscript{\rm 1}Nankai University, Tianjin, China; 
    \textsuperscript{\rm 2}Zhongguancun Academy, Beijing, China\\
    \textsuperscript{3} Beihang University, Beijing, China\\
    \Letter zwzhang@buaa.edu.cn
}

\begin{document}
\nocopyright
\maketitle

\begin{abstract}
Agents for long term reasoning require a memory that can be efficiently and effectively updated over time, as new facts and external feedback continue to arrive. Recently, graph memory has been adopted to offer structural organization for multi-hop retrieval and reasoning. However, existing methods store all memories in a flat graph, and accumulated historical memories can introduce irrelevant contexts and increase the cost of evidence selection during retrieval. Moreover, they typically update memory units independently, requiring repeated unit-wise rewrite to cover related changes. To address these issues, we propose \textsc{\methodname}, an evolving hierarchical graph memory framework with path-level localization and rewriting. Specifically, we first propose a hierarchical graph memory, which organizes the memory into coarse-to-fine architecture composed of upper-level nodes and MemoryUnits, thereby reducing the amount of irrelevant information during retrieval. We further propose MicroGraph-based path-level localization, which leverages query and update conditioned MicroGraphs to identify support subgraph and evidence path before rewrite. Finally, we propose a coordinated rewriting method that jointly revises intra-unit memory and inter-unit dependencies, enable valid dependency structures updating in the localized evidence path. Experiments on benchmarks for long-term conversational question answering and conflict-aware memory evaluation demonstrate that our method demonstrate substantial improvements over baselines in answer quality and token efficiency. Besides, our method improves answer accuracy and query-valid evidence selection under dynamic, static, and conditional conflicts.

\end{abstract}

\section{Introduction}

Long-term reasoning agents require memory mechanisms that can be efficiently and effectively updated over time as new facts, corrections, and external feedback continue to arrive. Recent memory-augmented systems improve long-horizon interaction by storing conversation histories, retrieving relevant experiences, compressing previous contexts, or maintaining personalized memory stores~\citep{zhong2024memorybank,packer2023memgpt,lee2024readagent,chhikara2025mem0}. However, continuously evolving memory introduces new challenges. The system should not only retrieve relevant historical information but also maintain its memory structure appropriately, thereby ensuring efficient access to reliable evidence for subsequent reasoning.

Graph-based memory has recently been adopted to provide structural organization for entities, relations, events, and temporal information, supporting multi-hop retrieval and reasoning~\citep{xu2026mem,rasmussen2025zep,chhikara2025mem0}. However, existing graph memory approaches often lack explicit coarse-to-fine organization for efficiently locating query-relevant evidence regions during memory maintenance. As accumulated historical memories grow, retrieval over this flat structure may introduce substantial irrelevant context, thereby increasing the cost of evidence selection. Moreover, these methods typically update memory units independently. Since answers are generally supported by interconnected evidence paths rather than isolated facts, independent unit-wise updates may omit relevant evidence and allow outdated dependencies to remain involved in subsequent reasoning. Repeated rewriting is therefore required to cover all related changes, leading to unbearable token consumption and low update efficiency.

To tackle these issues, we argue that the existing methods suffer from a mismatch between the granularity of memory organization and updates and that of the evidence structures used for reasoning. Retrieval operates over an continuously expanding overall graph, whereas an answer typically depends on only a small amount of localized evidence. Although an update may target a single memory unit, its effects can propagate along evidence paths. The memory system should therefore first localize the query-relevant subgraph, then identify the affected evidence paths based on the update, and jointly revise memory states and their dependencies within the bounded region. This process reduces irrelevant retrieval and repeated rewriting.

Based on this motivation, we propose \textsc{\methodname}, an evolving hierarchical graph memory framework with path-level localization and rewriting. Specifically, we first propose a hierarchical graph memory architecture with a coarse-to-fine structure composed of upper-level nodes and MemoryUnits. The upper-level nodes represent abstractions for MemoryUnits according to their subjects, object categories, and contexts, while MemoryUnits preserve fine-grained factual information and explicit dependencies. This organization reduces the amount of irrelevant information during retrieval without traversing the entire graph memory.
We further propose MicroGraph-based path-level localization for reasoning and rewriting. Given a query and an update, we first construct temporary MemoryUnits and extract anchors to locate relevant MicroGraphs and construct a localized support subgraph. We then identify evidence path affected by both the current query and update, determining the rewrite path before any memory rewrite.
Finally, we propose a coordinated rewriting method that jointly revises intra-unit memory and inter-unit dependencies within the localized evidence path. The intra-unit rewriting updates the internal states of affected MemoryUnits according to the new update, while the inter-unit rewriting rewrites dependency structures to maintain valid evidence connections. By coordinating memory state updates and dependency rewrites, the framework enables memory evolution while avoiding repeated unit-level modifications.

We evaluate \textsc{\methodname} on benchmarks for long-term conversational question answering LoCoMo~\citep{maharana2024evaluating} and conflict-aware memory evaluation MemConflict~\citep{tao2026memconflict}. 

Experimental results demonstrate that our method achieves substantial improvements over strong baselines in answer quality and token efficiency. Furthermore, our method improves answer accuracy and query-valid evidence selection under dynamic, static, and conditional conflicts. Our contributions are as follows:
\begin{itemize}
    \item We propose \textsc{\methodname}, a hierarchical graph memory framework, which organizes memory into coarse-to-fine structures, reducing irrelevant context and localizaiton cost caused by accumulated historical memories.
    \item We further propose a MicroGraph-based path-level localization method, which identifies a support subgraph and affected evidence path, thereby determining a narrowed explicit rewrite region.
    \item We also design a coordinated rewriting method that jointly revises the internal states of MemoryUnits and their dependencies within the localized region. 
\end{itemize}

\begin{figure*}[ht!]
\begin{center}
      \includegraphics[width=\textwidth]{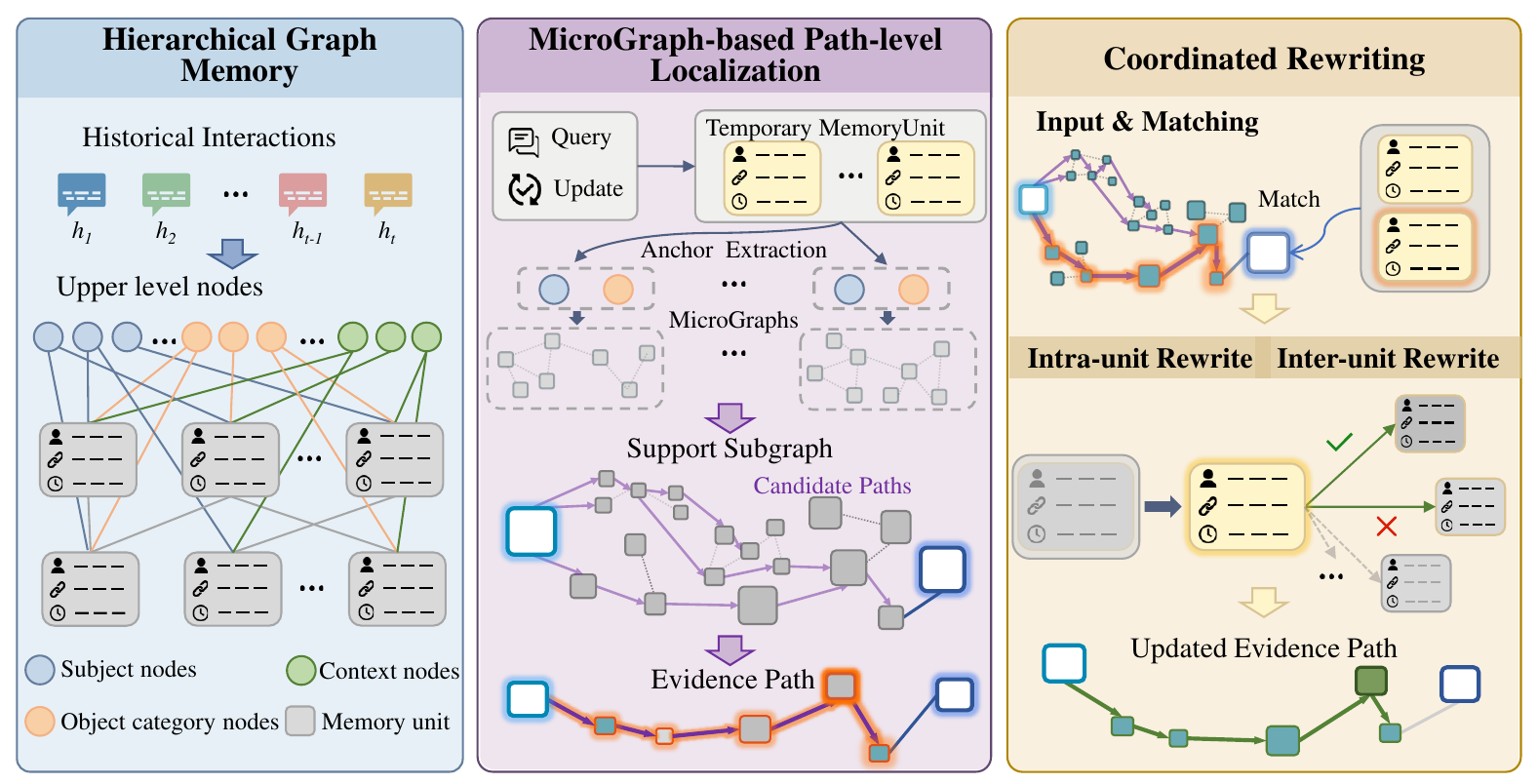}
\end{center}
  \caption{An overview of \textsc{\methodname}. Our method organizes memory into a hierarchical graph structure with upper-level nodes connecting MemoryUnits that store factual information. Then, given a query and an update, our MicroGraph-based path-level localization module retrieves relevant memory MicroGraphs to build a support subgraph, and identifies the affected evidence path. Lastly, coordinated rewriting updates MemoryUnit states and dependency structures within the localized evidence path to maintain consistent memory updates.}
    \label{fig:framework}
\end{figure*}

\section{Related Work}

\paragraph{Long-Term and Structured Memory.}
Long-term memory has become essential for LLM agents operating across multiple sessions.
Early approaches store dialogue histories, reusable experiences, or compressed memories~\citep{xu2022longtime,lu2023memochat,zhong2024memorybank,wang2023longmem}.
Later studies improve memory scalability through hierarchical management, virtual context expansion, and latent memory mechanisms~\citep{packer2023memgpt,wang2023longmem,lee2024readagent,wang2024memoryllm}, with recent methods further exploring multi-level organization for long-horizon reasoning~\citep{hmem2026,hierarchical_graph_memory2026,gmemory2025,kang2025memoryos,li2025memos}.
Agent-oriented frameworks incorporate reflection, self-improvement, and evolving user profiles to support long-term adaptation and personalization~\citep{park2023generative,shinn2023reflexion,yao2023react,liu2023thinkinmemory,yuan2023personalized}

Structured memory methods further explore graph-based and temporal architectures to organize entities, relations, and historical information~\citep{xu2026mem,chhikara2025mem0,rasmussen2025zep,gutierrez2024hipporag,anokhin2024arigraph,edge2024graphrag}.
As memory grows, operating on the entire memory graph introduces irrelevant information and increases evidence localization costs.
Few recent and concurrent works adopt hierarchical graph memory to organize memories through abstraction or evolution~\citep{hierarchical_graph_memory2026,gmemory2025}, but mainly focus on memory consolidation rather than coarse-to-fine localization of graph-based evidence structures.
In comparison, \methodname~organizes MemoryUnits into coarse-grained regions through a hierarchical memory architecture for efficient memory access.

\paragraph{Evidence Localization for Reasoning and Memory Maintenance.}

Recent memory systems study how to organize, retrieve, and update stored information to support long-term agent adaptation~\citep{zhong2024memorybank,packer2023memgpt,xu2026mem,chhikara2025mem0,kang2025memoryos,li2025memos}.
Structured memory approaches further maintain temporal and personalized information through graph-based representations~\citep{rasmussen2025zep}.
Recent studies show that effective reasoning requires query-conditioned evidence localization rather than simply expanding context~\citep{karpukhin2020dpr,guu2020realm,liu2024lost,trivedi2022ircot,edge2024graphrag,guo2024lightrag,li2024graphreader,sarthi2024raptor,yao2023tree,besta2024graph}.
However, existing memory maintenance methods mainly focus on storing, linking, or revising individual memory units, without explicitly identifying affected evidence structures before modification.
In evolving memory systems, new updates may influence multiple dependent memories beyond directly matched information.
\methodname~addresses this limitation by introducing path-level localization, which jointly considers query- and update-conditioned evidence to identify affected evidence paths and determine the rewrite region before memory revision.

\paragraph{Continual Memory Update and Conflict-Aware Revision.}
Maintaining memory consistency under evolving information requires modeling temporal validity, conflicts, and historical changes.
Temporal knowledge graph methods capture evolving facts through temporal representations~\citep{allen1983maintaining,trivedi2017knowevolve,cai2023temporal,qin2021timedial}, while recent memory systems study lifecycle management including storage, update, consolidation, and preservation~\citep{packer2023memgpt,zhong2024memorybank,wang2023longmem,wang2024memoryllm,chhikara2025mem0,rasmussen2025zep,kang2025memoryos}.
Continual knowledge revision and conflict-aware memory studies further investigate incorporating new information while maintaining consistency~\citep{meng2022rome,meng2023memit,chen2024lifelong,mitchell2022serac,xu2024knowledgeconflicts,wang2023resolving,pham2024whoswho}.
However, existing approaches mainly revise individual memory units or isolated facts, requiring repeated global searches to identify relevant memories for each update. 
\methodname~addresses this limitation through MicroGraph-based localization and coordinated rewriting, which jointly updates MemoryUnit states and inter-unit dependencies within the localized evidence region.

\section{The Proposed Method}
As illustrated in Figure~\ref{fig:framework}, \textsc{\methodname} maintains an hierarchical graph memory and performs memory localization and rewriting through three stages.
First, the hierarchical graph memory organization builds an abstraction structure over MemoryUnits, enabling coarse-grained access to relevant memory.
Second, MicroGraph-based path-level localization narrows the search space and identifies the path affected by the query and update before rewrite.
Lastly, coordinated rewriting updates MemoryUnit states and their dependencies within the localized region. 

\subsection{Hierarchical Memory Organization}


We represent the memory at $t$ as a hierarchical graph memory
$G_t=(\mathcal{V}_t,\mathcal{E}_t)$.
The node set is defined as
\begin{equation}
    \mathcal{V}_t
    =
    \mathcal{V}_t^{\mathrm{sub}}
    \cup
    \mathcal{V}_t^{\mathrm{cat}}
    \cup
    \mathcal{V}_t^{\mathrm{ctx}}
    \cup
    \mathcal{M}_t ,
\end{equation}
where $\mathcal{V}_t^{\mathrm{sub}}$,
$\mathcal{V}_t^{\mathrm{cat}}$, and
$\mathcal{V}_t^{\mathrm{ctx}}$ denote subject nodes, object-category nodes, and
context nodes, which serve as upper-level nodes, and $\mathcal{M}_t$ denotes the set of MemoryUnits that store fact. The upper-level nodes organize factual memories according to their roles and form an abstraction layer, which make the localization and rewriting more efficient.

Then, the edge set of the hierarchical graph memory is defined as
\begin{equation}
    \mathcal{E}_t
    =
    \mathcal{E}_t^{\mathrm{sub}}
    \cup
    \mathcal{E}_t^{\mathrm{obj}}
    \cup
    \mathcal{E}_t^{\mathrm{ctx}}
    \cup
    \mathcal{E}_t^{\mathrm{dep}} ,
\end{equation}
where $\mathcal{E}_t^{\mathrm{sub}} \subseteq \mathcal{V}_t^{\mathrm{sub}} \times \mathcal{M}_t$, $\mathcal{E}_t^{\mathrm{obj}} \subseteq \mathcal{V}_t^{\mathrm{obj}} \times \mathcal{M}_t $ and $\mathcal{E}_t^{\mathrm{ctx}} \subseteq \mathcal{V}_t^{\mathrm{ctx}} \times \mathcal{M}_t$ denotes the connections between each MemoryUnit with its corresponding
subject, object category, and contextual information, and $\mathcal{E}_t^{\mathrm{dep}} \subseteq \mathcal{M}_t \times \mathcal{M}_t$ denotes the dependency edges that connect MemoryUnits, representing explicit evidence dependencies.

Each MemoryUnit represents an independently retrievable and editable fact.
It records detailed information including the subject, relation, object, object category,
transaction time, context, confidence, current status, etc.
The status of a MemoryUnit indicates whether it is active, superseded,
outdated, or pending. Active MemoryUnits participate in current evidence retrieval.
Superseded and pending units preserve historical assertions after memory
evolution, while outdated units indicate dependent evidence whose supporting
information is no longer valid.
These non-active units remain in the graph to maintain revision history but are excluded from the localization step.

This hierarchical design separates organization from factual storage. 
Upper-level nodes provide coarse access, while MemoryUnits store fine-grained information and dependency structures.
Compared with flat graph memory that directly searches over all historical
facts, our organization enable identifying relevant regions and then
accesses corresponding MemoryUnits.
This design reduces unnecessary localization over accumulated memories and
provides a structured basis for subsequent evidence localization.

\subsection{MicroGraph-based Path-Level Localization}

\paragraph{MicroGraph construction}
Based on the hierarchical graph memory, we organize memory into MicroGraphs as localized regions rather than additional memory layers during localization.
Specifically, a MicroGraph $B_{t,j} \subseteq G_t$ is a subgraph of the global hierarchical graph memory determined by a pair of subject node and object-category node, the subject node identifies the entity-centered memory region, while the object-category node provides a coarse semantic constraint over stored facts. We choose these two attributes because they are stable across temporal updates and provide efficient access before detailed evidence path selection which jointly define a localized region.
Each MicroGraph corresponds to a pair
of nodes:
\begin{equation}
    B_{t,j}=(v_{t,j}^{\mathrm{sub}},v_{t,j}^{\mathrm{cat}}),
    \quad
    v_{t,j}^{\mathrm{sub}}\in\mathcal{V}_t^{\mathrm{sub}},
    \quad
    v_{t,j}^{\mathrm{cat}}\in\mathcal{V}_t^{\mathrm{cat}},
\end{equation}
and the associated MemoryUnits are defined as
\begin{equation}
    \mathcal{M}_t(B_{t,j})
    =
    \left\{
        m_i\in\mathcal{M}_t
        \mid
        s_i=v_{t,j}^{\mathrm{sub}},
        o_i=v_{t,j}^{\mathrm{cat}}
    \right\},
\end{equation}
where $m_i$ is a MemoryUnit, $s_i$ is its subject node,
and $o_i$ is its object-category node.
A MicroGraph does not introduce additional memory content.
Instead, it represents a localized region of the global graph memory that
contains the corresponding MemoryUnits, context nodes, and dependency edges,
enabling efficient retrieval and evidence localization.

Given a query $q_t$ and an available update text $u_t$, we first construct
temporary MemoryUnits $\mathcal{M}_t^{\mathrm{temp}}$.
These units follow the MemoryUnit schema but remain outside the
graph memory during localization.
Only update-derived units are committed after rewriting.

The anchor extractor obtains the subject nodes and object-category nodes
from temporary MemoryUnits:
\begin{equation}
    \mathcal{A}_t
    =
    \operatorname{anch}
    \left(
        \mathcal{M}_t^{\mathrm{temp}}
    \right),
\end{equation}
where $\operatorname{anch}(\cdot)$ extracts the anchors used for
MicroGraph localization. 
The extracted anchors are used to identify relevant MicroGraphs.
The candidate MicroGraphs are defined as
\begin{equation}
\resizebox{0.95\columnwidth}{!}{
$
\displaystyle
\mathcal{B}_t^{\mathrm{cand}}
=
\left\{
B\in\mathcal{B}_t
\;\middle|\;
\exists m_i\in\mathcal{M}_t(B):
\{s_i,o_i\}
\cap
\mathcal{A}_t
\neq
\varnothing
\right\},
$
}
\end{equation}
where $\mathcal{M}_t(B)$ denotes the MemoryUnits associated with
MicroGraph $B$.
The candidate MicroGraphs are then ranked according to their relevance to the
extracted anchors.
We denote the relevance score as $R(B,\mathcal{A}_t)$, which measures the
subject matching and object-category compatibility between the anchors and
the MemoryUnits associated with each MicroGraph.
The top-$K_g$ MicroGraphs are selected as the localized memory region:
\begin{equation}
\widehat{\mathcal{B}}_t
=
\operatorname{TopK}_{K_g}
\left(
\mathcal{B}_t^{\mathrm{cand}},
R
\right).
\end{equation}
The MemoryUnits associated with $\widehat{\mathcal{B}}_t$, together with
their subject nodes, object-category nodes, context nodes, and
dependency edges, form the support subgraph $G_{S,t}$.

\paragraph{Path-level localization.}
Although the support subgraph narrows the search space, it still contains
multiple evidence structures. Since updates may affect only specific evidence
paths, we further perform path-level localization to identify the evidence
path for subsequent rewriting.
An evidence path represents a connected chain within the localized
subgraph, where the involved MemoryUnits and dependency edges collectively
support the answer to the current query.
Within the localized evidence subgraph $G_{S,t}$, we enumerate connected
MemoryUnit paths for evidence selection and update-impact analysis.
Adjacent MemoryUnits are connected through explicit dependency edges or valid
structural relations retained in $G_{S,t}$.
We enumerate up to $K_p$ candidate paths in $G_{S,t}$ by following these
connections, denoted as $\mathcal{P}_t^{\mathrm{cand}}$.

Each candidate path is further evaluated according to its relevance to the
temporary MemoryUnits, denoted as 
$
    \boldsymbol{\phi}_H
    \left(
        P,\mathcal{M}_t^{\mathrm{temp}}
    \right),
$,
which captures the consistency between a candidate path and the temporary MemoryUnits.
The scoring function considers the matching of MemoryUnit attributes,
dependency consistency, temporal validity, and contextual compatibility.
The affected evidence path is selected as
\begin{equation}
    \widehat{P}_t
    =
    \arg\max \nolimits_{P\in\mathcal{P}_t^{\mathrm{cand}}}
    \boldsymbol{\phi}_H
    \left(
        P,\mathcal{M}_t^{\mathrm{temp}}
    \right)).
\end{equation}
The MemoryUnits in $\widehat{P}_t$ and their associated
dependency edges define the rewrite region for coordinated rewriting.

\begin{table*}[t]
\centering
\begingroup
\setlength{\tabcolsep}{1.2pt}
\renewcommand{\arraystretch}{0.92}
\fontsize{7.0}{9.0}\selectfont

\newcommand{\benchVCell}[1]{%
    \rotatebox[origin=c]{90}{#1}%
}

\resizebox{\linewidth}{!}{%
\begin{tabular}{
@{}
cc
@{\hspace{1.4pt}}ccc
@{\hspace{1.4pt}}ccc
@{\hspace{1.4pt}}ccc
@{\hspace{1.4pt}}ccc
@{\hspace{1.4pt}}ccc
@{\hspace{1.4pt}}ccc
@{\hspace{1.4pt}}r
@{}
}
\toprule


\multirow[c]{2}{*}{LLM}
& \multirow[c]{2}{*}{Method}
& \multicolumn{3}{c}{Single Hop}
& \multicolumn{3}{c}{Multi-Hop}
& \multicolumn{3}{c}{Open Domain}
& \multicolumn{3}{c}{Temporal}
& \multicolumn{3}{c}{Adversarial}
& \multicolumn{3}{c}{Average}
& \multirow[c]{2}{*}{\shortstack{Token\\Length}}
\\




\cmidrule(lr){3-5}
\cmidrule(lr){6-8}
\cmidrule(lr){9-11}
\cmidrule(lr){12-14}
\cmidrule(lr){15-17}
\cmidrule(lr){18-20}

&
& F1 & BLEU & LLM-J
& F1 & BLEU & LLM-J
& F1 & BLEU & LLM-J
& F1 & BLEU & LLM-J
& F1 & BLEU & LLM-J
& F1 & BLEU & LLM-J
&
\\

\midrule


\multirow[c]{7}{*}{\benchVCell{GPT-5.4}}
& LoCoMo
& \underline{63.43} & \textbf{59.13} & \underline{83.98}
& \textbf{47.80} & \textbf{39.15} & \textbf{78.72}
& \textbf{22.85} & \textbf{19.77} & \textbf{64.58}
& \underline{30.88} & \underline{24.91} & 59.19
& 74.89 & 74.89 & 88.34
& \underline{47.97} & \underline{43.57} & \underline{74.96}
& 28345.6
\\

& MemoryBank
& 45.87 & 42.62 & 77.88
& 21.60 & 15.65 & 57.09
& 13.29 & 12.17 & 59.38
& 20.99 & 16.14 & 66.98
& \underline{82.51} & \underline{82.51} & 88.83
& 36.85 & 33.82 & 70.03
& 820.5
\\

& A-MEM
& 51.66 & 48.32 & 81.33
& 22.75 & 16.21 & 59.93
& 9.18 & 6.92 & 60.42
& 23.50 & 18.25 & 64.49
& 79.37 & 79.37 & \underline{89.69}
& 37.29 & 33.81 & 71.17
& \underline{586.8}
\\

& ReadAgent
& 59.34 & 53.38 & 82.74
& 38.72 & 36.61 & 68.05
& \underline{22.20} & \underline{18.64} & \underline{62.50}
& 30.00 & 24.47 & 59.19
& 67.25 & 67.25 & 87.00
& 43.50 & 40.07 & 71.90
& 14679.7
\\

& MemGPT
& 20.47 & 18.69 & 60.64
& 19.02 & 13.76 & 56.03
& 8.87 & 7.35 & 57.29
& 11.12 & 9.44 & \underline{74.14}
& 81.53 & 81.53 & 85.60
& 28.20 & 26.15 & 66.74
& 4180.3
\\

& Mem0
& 51.26 & 47.69 & 80.98
& 22.95 & 16.23 & 61.35
& 9.90 & 7.93 & 60.42
& 27.49 & 21.90 & 62.62
& 53.59 & 53.59 & 79.82
& 33.04 & 29.47 & 69.04
& \textbf{517.8}
\\


& \textbf{\methodname}
& \textbf{63.96} & \underline{55.74} & \textbf{85.85}
& \underline{44.27} & \underline{37.01} & \underline{69.50}
& 19.64 & 16.16 & 61.46
& \textbf{42.05} & \textbf{38.69} & \textbf{86.92}
& \textbf{83.18} & \textbf{83.18} & \textbf{90.44}
& \textbf{50.62} & \textbf{46.16} & \textbf{78.83}
& 2031.5
\\


\cmidrule(lr){1-21}

\multirow[c]{7}{*}{\benchVCell{GPT-4o}}
& LoCoMo
& \underline{64.67} & \underline{59.63} & \textbf{91.91}
& \underline{44.39} & \underline{36.92} & \textbf{80.50}
& \textbf{16.72} & \underline{14.35} & \textbf{65.62}
& \underline{38.78} & \underline{33.71} & 61.68
& 87.89 & 87.89 & 94.39
& \underline{50.49} & \underline{46.50} & 78.82
& 28346.5
\\

& MemoryBank
& 40.26 & 37.38 & 79.67
& 16.03 & 11.80 & 68.79
& 6.65 & 5.81 & 63.54
& 31.80 & 28.65 & 81.00
& 87.64 & 86.64 & 88.43
& 36.48 & 34.06 & 76.29
& 819.9
\\

& A-MEM
& 44.70 & 41.62 & 81.45
& 16.61 & 12.51 & 69.15
& 4.10 & 3.53 & 61.46
& 33.26 & 29.38 & 77.57
& \textbf{97.31} & \textbf{97.31} & \textbf{98.43}
& 39.20 & 36.87 & 77.61
& \underline{586.3}
\\

& ReadAgent
& 63.25 & 58.44 & \underline{91.08}
& 41.82 & 33.68 & 72.70
& 10.02 & 7.72 & 64.62
& 35.43 & 31.29 & 73.52
& 87.44 & 87.44 & 95.96
& 47.59 & 43.71 & \underline{79.58}
& 14723.1
\\

& MemGPT
& 17.17 & 15.81 & 70.39
& 16.01 & 12.40 & 68.44
& 6.04 & 5.01 & 61.46
& 10.32 & 8.96 & \underline{83.18}
& 92.41 & 92.46 & 91.65
& 28.39 & 26.93 & 75.02
& 4178.5
\\

& Mem0
& 37.90 & 35.49 & 77.29
& 11.30 & 8.07 & 68.09
& 2.47 & 2.18 & 60.42
& 26.82 & 23.34 & 77.26
& 81.39 & 81.39 & 93.72
& 31.98 & 30.09 & 75.36
& \textbf{516.5}
\\


& \textbf{\methodname}
& \textbf{65.41} & \textbf{61.76} & 87.16
& \textbf{47.96} & \textbf{42.45} & \underline{78.82}
& \underline{15.46} & \textbf{14.67} & \underline{65.50}
& \textbf{38.87} & \textbf{35.43} & \textbf{88.79}
& \underline{94.84} & \underline{94.84} & \underline{97.98}
& \textbf{52.51} & \textbf{49.83} & \textbf{83.65}
& 2328.9
\\

\bottomrule

\end{tabular}%
}
\caption{Performance comparison of different question categories on LoCoMo. The best results are shown in \textbf{Bold} and the second-best results are \underline{underlined}.}
\label{tab:main}
\endgroup
\end{table*}

\subsection{Coordinated Rewriting}

The goal of coordinated rewriting is not to overwrite historical memories, but to maintain a consistent evidence structure where updated facts and their dependent conclusions remain valid. After path-level localization identifies
$\widehat{P}_t$, we perform coordinated rewriting
within the fixed evidence region. The temporary MemoryUnits
$\mathcal{M}_t^{\mathrm{temp}}$ are first matched with the MemoryUnits in
$\widehat{P}_t$ to identify the affected MemoryUnits. 
Then, the intra-unit rewriting updates the internal states of these MemoryUnits based
on the temporary updates, while the status of updated MemoryUnits are set as active because they represent the current valid evidence.
The state changes of updated MemoryUnits further determine the affected
dependency relations. Inter-unit rewriting next revises the dependency
structures connected to the affected MemoryUnits and examines whether the
related MemoryUnits require additional updates according to
$\mathcal{M}_t^{\mathrm{temp}}$. Both rewriting operations are performed
within the same localized evidence region and are updated together:
\begin{equation}
\resizebox{0.99\linewidth}{!}{$\displaystyle
G_{t+1}
=
\operatorname{Update}\Big(
G_t,
\operatorname{Rewrite}_{\mathrm{inter}}
\Big(
\operatorname{Rewrite}_{\mathrm{intra}}
(
\widehat{P}_t,
\mathcal{M}_t^{\mathrm{temp}}
),
\mathcal{M}_t^{\mathrm{temp}}
\Big)
\Big).
$}
\label{eq:coordinated_rewrite}
\end{equation}
where $\operatorname{Rewrite}_{\mathrm{intra}}(\cdot)$ updates the internal states of
affected MemoryUnits and $\operatorname{Rewrite}_{\mathrm{inter}}(\cdot)$ revises
the dependency relations according to the updated MemoryUnits. This
coordination enables memory states and dependency structures to evolve with
respect to the same update evidence. MemoryUnits and relations outside the
rewrite region remain unchanged. The detailed rewriting methods are as follows.

\paragraph{Intra-unit rewriting.}
For each temporary MemoryUnit
$\bar{m}_j\in\mathcal{M}_t^{\mathrm{temp}}$, we identify the matched
MemoryUnit $m_i$ within the affected path
$\widehat{P}_t$.
The rewriting process preserves existing records and updates the internal
state and valid-time information of affected MemoryUnits according to the
new evidence.
If $\bar{m}_j$ introduces a new memory, it is committed as an active
MemoryUnit. If it updates an existing memory, the corresponding
MemoryUnit state is revised according to the temporal and contextual
consistency with $\bar{m}_j$.

\paragraph{Inter-unit rewriting.}
The state revision of an affected MemoryUnit may influence other MemoryUnits
whose evidence depends on it. Our method identifies the directly dependent
MemoryUnits within the evidence path:
\begin{equation}
\mathcal{D}_t(m_i)
=
\left\{
m\in\mathcal{M}_t(\widehat{P}_t)
\;\middle|\;
(m_i,m)\in\mathcal{E}_t^{\mathrm{dep}}
\right\},
\label{eq:dependent_units}
\end{equation}
where $\mathcal{M}_t(\widehat{P}_t)$ denotes the MemoryUnits
contained in the affected evidence path. Dependency edges are directed from
supporting MemoryUnits to dependent MemoryUnits.

After intra-unit rewriting, we examine whether the dependencies of
each MemoryUnit in $\mathcal{D}_t(m_i)$ remain valid under the updated
evidence. A dependent MemoryUnit is preserved when its supporting evidence
remains consistent. Otherwise, it is marked as outdated and excluded from the
current evidence view.
The dependencies of an updated MemoryUnit are not directly inherited from the
original MemoryUnit. The updated MemoryUnit may have a different evidence scope and thus cannot
always support the original downstream conclusions. Dependencies involving the updated MemoryUnit are retained only
when supported by valid evidence, preventing unsupported derived conclusions.
\section{Experiments}
\label{sec:experiments}

We evaluate \methodname{} by answering the following four research questions:
\begin{itemize}
\item \textbf{RQ1:} Can \methodname~improve long-term QA quality and token efficiency?
\item \textbf{RQ2:} Can \methodname~maintain consistent memory under different conflicts?
\item \textbf{RQ3:} How do memory organization and evidence localization affect performance?
\item \textbf{RQ4:} How does \methodname~compare with different memory update strategies?
\end{itemize}

\subsection{Experimental Setup}

\paragraph{Datasets and Evaluation Metrics.}
We evaluate \methodname~on two commonly used benchmarks for agent memory. 

\textbf{LoCoMo}~\cite{maharana2024evaluating} contains long-term conversations together with questions and reference answers derived from them. The questions are divided into five categories: Single Hop, Multi-Hop, Open Domain, Temporal, and Adversarial. For each category, we report token-level F1, BLEU-1 (BLEU), and the LLM-as-Judge score (LLM-J). 

F1 and BLEU measure lexical overlap between the generated and reference answers, whereas LLM-J evaluates semantic correctness. Besides, Token Length reports the total number of input and output tokens used by the final answer generation call, excluding offline memory construction and update. 

\begin{figure*}[t]
    \centering
    \includegraphics[width=\textwidth]{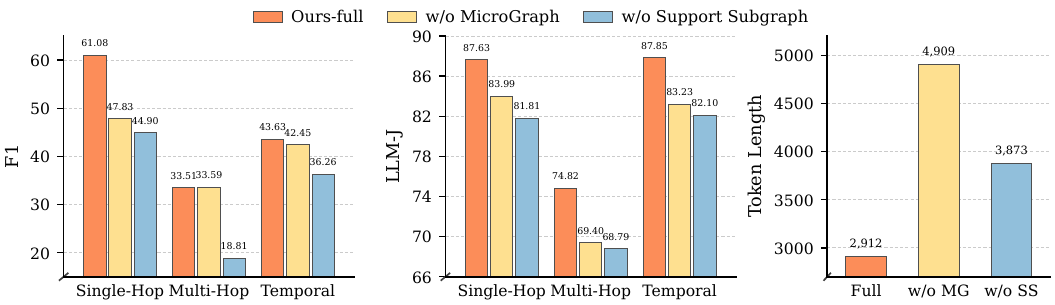}
    \caption{
       Results of ablation of Memory Organization and Evidence Localization on LoCoMo. 
       \emph{w/o MicroGraph} removes MicroGraph organization. \emph{w/o Support Subgraph} removes the localized support region. 
    }
    \label{fig:ablation}
\end{figure*}

\textbf{MemConflict}~\cite{tao2026memconflict} contains long interaction histories with conflicts divided by three categories. The Dynamic subset tests temporal validity. The Static subset measures resistance to incorrect replacement. The Conditional (Cond) subset tests whether a value applies under the current condition. Following the benchmark, we adopt different metrics for different categories including Answer Accuracy (AA), Update Order Consistency Score (UOCS) which measures consistency under update order for dynamic conflicts, Conflict Recognition Score (CRS) which measures recognition of static conflicts, Macro Answer Accuracy (Macro-AA) which is the unweighted average over the three conflict types, SEH@3 which records whether the gold evidence appears among the top three retrieved items, and Support Rank Score (SRS) which applies a logarithmic discount to the evidence rank. 
Higher values are better for every metric.

\paragraph{Baselines.}
On LoCoMo, we compare with six representative baselines. \textbf{LoCoMo}~\cite{maharana2024evaluating} places the entire dialogue history in the generation prompt. \textbf{MemoryBank}~\cite{zhong2024memorybank} stores textual memories and adjusts retention using a forgetting-curve mechanism. \textbf{A-MEM}~\cite{xu2026mem} constructs atomic notes with contextual descriptions, keywords, and tags, dynamically updates their attributes. \textbf{ReadAgent}~\cite{lee2024readagent} compresses episodes into gist memories and revisits selected pages on demand. \textbf{MemGPT}~\cite{packer2023memgpt} manages state across context and archival memory tiers. \textbf{Mem0}~\cite{chhikara2025mem0} extracts salient facts and applies fact-level  operations after retrieving related memories. On MemConflict, we evaluate A-MEM and Mem0 together with three additional methods. \textbf{LangMem}~\cite{langmem2026software} extracts, consolidates, and updates semantic memories. \textbf{Letta}~\cite{letta2026software} maintains persistent agent state through managed memory tiers. \textbf{MemOS}~\cite{li2025memos} manages memory representation, storage, retrieval, and lifecycle. 

\paragraph{Implementation Details.}
We use GPT 5.4 and GPT 4o to generate answers on LoCoMo while GPT 5.0 mini is used on MemConflict. GPT 5.4 mini serves as the judge model. The maximum output length is set as 256 tokens on LoCoMo and 4,096 tokens on MemConflict. The default path depth is $h = 3$. We set $K_g = 12$ and the evidence budget $K_p = 24$.

\begin{table}[t]
\centering

\setlength{\tabcolsep}{2.5pt}
\renewcommand{\arraystretch}{1.12}
\resizebox{\columnwidth}{!}{%
\begin{tabular}{lcccccccc}
\toprule
\multirow{2}{*}{Method} & \multicolumn{2}{c}{Dynamic} & \multicolumn{2}{c}{Static} & \multicolumn{1}{c}{Cond} & \multicolumn{3}{c}{Overall} \\
\cmidrule(lr){2-3}\cmidrule(lr){4-5}\cmidrule(lr){6-6}\cmidrule(lr){7-9}
& AA & UOCS & AA & CRS & AA & Macro-AA & SEH@3 & SRS \\
\midrule
MemOS   & 37.93 & \underline{38.18} & \underline{43.75} & 23.61 & \underline{84.49} & \underline{55.39} & \underline{67.10} & \underline{58.79} \\
LangMem & \textbf{49.66} & 35.79 & 19.44 & 20.83 & 15.56 & 28.22 & 43.49 & 39.10 \\
Letta   & 39.55 & 35.27 & 22.23 & 20.31 & 84.35 & 48.71 & 62.02 & 51.24 \\
Mem0    & 12.24 & 11.30 & 19.44 & 15.28 & 76.67 & 36.12 & 43.81 & 39.23 \\
A-MEM   & 35.96 & 29.11 & 26.39 & \underline{25.01} & 71.22 & 44.52 & 56.42 & 48.28 \\
\textbf{\methodname}    & \underline{44.78} & \textbf{49.14} & \textbf{68.75} & \textbf{68.06} & \textbf{90.00} & \textbf{67.84} & \textbf{81.06} & \textbf{77.31} \\
\bottomrule
\end{tabular}%
}
\caption{Performance comparison of different methods on the MemConflict benchmark. }\label{tab:memconflict-results}
\end{table}

\subsection{Main Results}
To answer RQ1, we first report the results on LoCoMo in Table~\ref{tab:main}. The results show that \methodname~achieves the best average F1, BLEU, and LLM-J under both GPT-5.4 and GPT-4o. This consistency across two LLM models suggests that the improvements are not tied to a particular LLM backbone. Instead, they reflect the ability of \methodname~to organize and select reliable evidence for answer generation.

\methodname~also provides a favorable balance between answer quality and inference cost. It uses only 7.2\% of the tokens consumed by full context under GPT-5.4 and 15.8\% of ReadAgent under GPT-4o, which achieves the second best results, respectively. Meanwhile, though several compact memory systems such as MemoryBank, A-MEM, and Mem0 require fewer tokens, their average answer quality is substantially lower. \methodname~reduces irrelevant historical context without discarding the evidence required for accurate generation. 

The results for different categories further reveal the advantages of our method. \methodname~achieves the strongest performance for Temporal questions under both answer models. It also ranks the first on Adversarial questions under GPT-5.4 and remains close to the best baseline under GPT-4o. These results are consistent with the use of temporal validity scopes and localized evidence paths, which limit the influence of outdated and unrelated memories. \methodname~remains competitive on Single-Hop and Multi-Hop questions, although the relative gains vary across answer models. Its results on Open Domain  are less dominant, which reflects the limitation of memory retrieval when the required external knowledge is absent from the stored history. Overall, the results demonstrate that selecting structured evidence is more effective than exposing the LLM to a large historical context.

To answer RQ2, we further report the results of MemConflict in Table~\ref{tab:memconflict-results}. The results show that \methodname~achieves the strongest overall performance and ranks first on both evidence selection metrics. The simultaneous improvements indicate that \methodname~not only produces more accurate answers, but also places valid supporting evidence earlier in the retrieval ranking. This distinction is important for MemConflict, where successful reasoning requires the system to identify evidence that remains applicable to the current query.

The results across conflict types reveal complementary strengths. On Dynamic conflicts, \methodname~does not achieve the highest AA, but obtains the best UOCS. This pattern indicates that its main advantage lies in maintaining consistency across different update orders. On Static conflicts, \methodname~leads both AA and CRS by a clear margin. This result is consistent with its rewrite design, which retains previous assertions and records their state transitions instead of overwriting memory directly. Dependency revision further prevents evidence supported by invalid states from remaining in the active view. \methodname~also achieves the best Conditional AA, suggesting that explicit contextual scopes help distinguish facts with different applicability conditions.

Overall, these results show that the advantage of \methodname~comes from coupling evidence localization with coordinated memory revision. Path level localization identifies evidence relevant to both the query and the update, while coordinated rewriting maintains the corresponding MemoryUnit states and dependencies within the localized region. The following analyses further isolate the contributions of memory organization, evidence localization, and rewriting.

\subsection{Analysis}
\paragraph{Ablations.}
To answer RQ3, we report the results of ablations of Memory Organization and Evidence Localization in  Figure~\ref{fig:ablation}. From the figure,  MicroGraph organization and localized evidence construction play complementary roles in retrieval. Removing MicroGraph organization increases token use by 68.6\% and reduces LLM-J across all three question categories. The substantial decline on Single-Hop questions shows that coarse-level indexing improves direct access to relevant evidence. Multi-Hop F1 remains nearly unchanged, while its LLM-J decreases. This suggests that a wider search space may preserve surface overlap but introduce evidence that is less reliable for reasoning.
Removing the localized support subgraph causes a broader decline across all three categories. The Single-Hop degradation shows that local structure also helps isolate the correct MemoryUnit for direct retrieval. The losses on Multi-Hop and Temporal questions further demonstrate the importance of preserving dependencies and temporal relations among retrieved units. Since token use only slightly increases, the performance drop is attributed to the loss of structured evidence before path scoring rather than insufficient context. 

\begin{table}[t]
\centering


\begin{tabular}{lccc}
\toprule
Variant
& Dynamic AA
& Static AA
& Average \\
\midrule
Append-only
& 42.72
& 43.06
& 42.89 \\
Relation-level
& 31.25
& 50.26
& 40.76 \\
\textbf{HiGram}
& \textbf{44.78}
& \textbf{68.75}
& \textbf{56.77} \\
\bottomrule
\end{tabular}
\caption{Comparison of different memory update strategies on MemConflict. Append-only represents insertion-based updates, Relation-level performs isolated updates.}
\label{tab:rewrite}
\end{table}

\paragraph{Memory Update Variants.}
To answer RQ4, we report the results of comparing different memory update methods in Table~\ref{tab:rewrite}. Append Only yields balanced results on dynamic and static conflicts, but remains limited in both settings. Relation-level Update improves static accuracy, yet its substantial decline on dynamic conflicts lowers its overall mean below Append Only. This contrast shows that improving one conflict type can weaken robustness across update settings. \methodname~avoids this trade-off by achieving the strongest accuracy on both subsets and the highest mean, with a particularly clear advantage on static conflicts. These consistent gains indicate that \methodname~improves conflict correction without sacrificing adaptation to evolving information.

\begin{figure}
    \centering
    \includegraphics[width=0.9\linewidth]{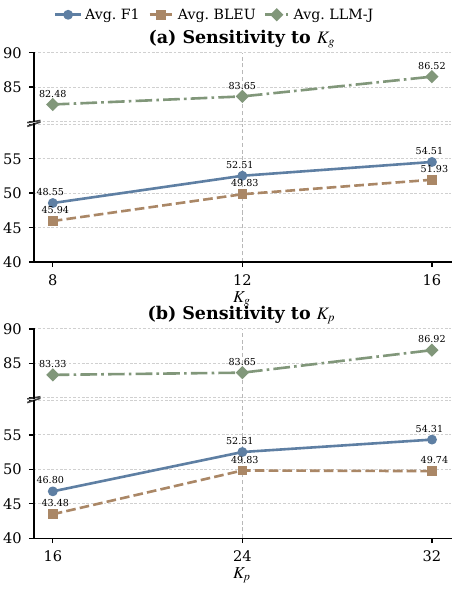}
    \caption{Sensitivity analysis of key hyperparameters on LoCoMo. 
$K_g$ denotes the number of retrieved MicroGraphs, and $K_p$ denotes the number of enumerate up evidence paths.}
    \label{fig:hyperparameter}
\end{figure}

\paragraph{Hyperparameter Sensitivity.}
Figure~\ref{fig:hyperparameter} evaluates the sensitivity of \methodname~to the number of retrieved MicroGraphs $K_g$ and candidate paths $K_p$. The results show that \methodname~maintains stable performance across different configurations. Increasing $K_g$ improves performance by providing more relevant memory regions, while the gains gradually saturate with larger retrieval scopes. Similarly, varying $K_p$ leads to limited performance changes, indicating that the path-level scoring mechanism can effectively identify relevant evidence paths under different search budgets. These results demonstrate that \methodname~is robust to hyperparameter variations and does not require careful tuning.
\section{Conclusion}
\label{sec:conclusion}

We present \methodname~, an evolving hierarchical graph memory framework for long-term reasoning agents. \methodname~organizes memory into coarse-to-fine localized evidence regions based on MicroGraph and uses path-level localization to identify evidence paths, together with a coordinated rewriting mechanism that jointly revises the internal states of MemoryUnits and their dependencies. Experiments on LoCoMo and MemConflict show that \methodname~improves answer quality and token efficiency. Future work will investigate external knowledge integration and multimodal memories.

\bibliography{refs}

@inproceedings{maharana2024evaluating,
  title={Evaluating very long-term conversational memory of llm agents},
  author={Maharana, Adyasha and Lee, Dong-Ho and Tulyakov, Sergey and Bansal, Mohit and Barbieri, Francesco and Fang, Yuwei},
  booktitle={Proceedings of the 62nd Annual Meeting of the Association for Computational Linguistics},
  pages={13851--13870},
  year={2024}
}

@inproceedings{zhong2024memorybank,
  title={Memorybank: Enhancing large language models with long-term memory},
  author={Zhong, Wanjun and Guo, Lianghong and Gao, Qiqi and Ye, He and Wang, Yanlin},
  booktitle={Proceedings of the AAAI conference on artificial intelligence},
  number={17},
  pages={19724--19731},
  year={2024}
}

@article{xu2026mem,
  title={A-mem: Agentic memory for llm agents},
  author={Xu, Wujiang and Liang, Zujie and Mei, Kai and Gao, Hang and Tan, Juntao and Zhang, Yongfeng},
  journal={Advances in Neural Information Processing Systems},
  volume={38},
  pages={17577--17604},
  year={2026}
}

@article{packer2023memgpt,
  title={MemGPT: towards LLMs as operating systems},
  author={Packer, Charles and Fang, Vivian and Patil, Shishir\_G and Lin, Kevin and Wooders, Sarah and Gonzalez, Joseph\_E},
  year={2023},
  journal={arXiv preprint arXiv:2310.08560},
}

@inproceedings{lee2024readagent,
  title={A Human-Inspired Reading Agent with Gist Memory of Very Long Contexts},
  author={Lee, Kuang-Huei and Chen, Xinyun and Furuta, Hiroki and Canny, John and Fischer, Ian},
  booktitle={International Conference on Machine Learning},
  pages={26396--26415},
  year={2024},
  organization={PMLR}
}

@article{chhikara2025mem0,
  title={Mem0: Building production-ready ai agents with scalable long-term memory},
  author={Chhikara, Prateek and Khant, Dev and Aryan, Saket and Singh, Taranjeet and Yadav, Deshraj},
  journal={arXiv preprint arXiv:2504.19413},
  year={2025}
}

@article{rasmussen2025zep,
  title={Zep: a temporal knowledge graph architecture for agent memory},
  author={Rasmussen, Preston and Paliychuk, Pavlo and Beauvais, Travis and Ryan, Jack and Chalef, Daniel},
  journal={arXiv preprint arXiv:2501.13956},
  year={2025}
}

@article{tao2026memconflict,
  title={MemConflict: Evaluating Long-Term Memory Systems Under Memory Conflicts},
  author={Tao, Zhen and Zhao, Jinxiang and Liu, Peng and Xi, Dinghao and Chen, Yanfang and Xu, Wei and Li, Zhiyu},
  journal={arXiv preprint arXiv:2605.20926},
  year={2026}
}

@article{wang2023longmem,
    title={Augmenting language models with long-term memory},
  author={Wang, Weizhi and Dong, Li and Cheng, Hao and Liu, Xiaodong and Yan, Xifeng and Gao, Jianfeng and Wei, Furu},
  journal={Advances in Neural Information Processing Systems},
  volume={36},
  pages={74530--74543},
  year={2023}
}

@inproceedings{wang2024memoryllm,
  title={MEMORYLLM: towards self-updatable large language models},
  author={Wang, Yu and Gao, Yifan and Chen, Xiusi and Jiang, Haoming and Li, Shiyang and Yang, Jingfeng and Yin, Qingyu and Li, Zheng and Li, Xian and Yin, Bing and others},
  booktitle={Proceedings of the 41st International Conference on Machine Learning},
  pages={50453--50466},
  year={2024}
}

@article{lu2023memochat,
  title={Memochat: Tuning llms to use memos for consistent long-range open-domain conversation},
  author={Lu, Junru and An, Siyu and Lin, Mingbao and Pergola, Gabriele and He, Yulan and Yin, Di and Sun, Xing and Wu, Yunsheng},
  journal={arXiv preprint arXiv:2308.08239},
  year={2023}
}

@inproceedings{xu2022longtime,
  title={Long time no see! open-domain conversation with long-term persona memory},
  author={Xu, Xinchao and Gou, Zhibin and Wu, Wenquan and Niu, Zheng-Yu and Wu, Hua and Wang, Haifeng and Wang, Shihang},
  booktitle={Findings of the Association for Computational Linguistics},
  pages={2639--2650},
  year={2022}
}

@inproceedings{park2023generative,
  title={Generative agents: Interactive simulacra of human behavior},
  author={Park, Joon Sung and O'Brien, Joseph and Cai, Carrie Jun and Morris, Meredith Ringel and Liang, Percy and Bernstein, Michael S},
  booktitle={Proceedings of the 36th annual acm symposium on user interface software and technology},
  pages={1--22},
  year={2023}
}

@article{shinn2023reflexion,
  title={Reflexion: Language agents with verbal reinforcement learning},
  author={Shinn, Noah and Cassano, Federico and Gopinath, Ashwin and Narasimhan, Karthik and Yao, Shunyu},
  journal={Advances in neural information processing systems},
  volume={36},
  pages={8634--8652},
  year={2023}
}

@article{liu2023thinkinmemory,
  title={Think-in-memory: Recalling and post-thinking enable llms with long-term memory},
  author={Liu, Lei and Yang, Xiaoyan and Shen, Yue and Hu, Binbin and Zhang, Zhiqiang and Gu, Jinjie and Zhang, Guannan},
  journal={arXiv preprint arXiv:2311.08719},
  year={2023}
}

@inproceedings{yuan2023personalized,
  title={Personalized large language model assistant with evolving conditional memory},
  author={Yuan, Ruifeng and Sun, Shichao and Li, Yongqi and Wang, Zili and Cao, Ziqiang and Li, Wenjie},
  booktitle={Proceedings of the 31st International Conference on Computational Linguistics},
  pages={3764--3777},
  year={2025}
}

@inproceedings{karpukhin2020dpr,
  title={Dense passage retrieval for open-domain question answering},
  author={Karpukhin, Vladimir and Oguz, Barlas and Min, Sewon and Lewis, Patrick and Wu, Ledell and Edunov, Sergey and Chen, Danqi and Yih, Wen-tau},
  booktitle={Proceedings of the 2020 conference on empirical methods in natural language processing},
  pages={6769--6781},
  year={2020}
}

@inproceedings{guu2020realm,
  title={Retrieval augmented language model pre-training},
  author={Guu, Kelvin and Lee, Kenton and Tung, Zora and Pasupat, Panupong and Chang, Mingwei},
  booktitle={International conference on machine learning},
  pages={3929--3938},
  year={2020}
}

@inproceedings{trivedi2022ircot,
  title={Interleaving retrieval with chain-of-thought reasoning for knowledge-intensive multi-step questions},
  author={Trivedi, Harsh and Balasubramanian, Niranjan and Khot, Tushar and Sabharwal, Ashish},
  booktitle={Proceedings of the 61st annual meeting of the association for computational linguistics},
  pages={10014--10037},
  year={2023}
}

@article{liu2024lost,
  title={Lost in the middle: How language models use long contexts},
  author={Liu, Nelson F and Lin, Kevin and Hewitt, John and Paranjape, Ashwin and Bevilacqua, Michele and Petroni, Fabio and Liang, Percy},
  journal={Transactions of the association for computational linguistics},
  volume={12},
  pages={157--173},
  year={2024}
}

@article{edge2024graphrag,
 title={From local to global: A graph rag approach to query-focused summarization},
  author={Edge, Darren and Trinh, Ha and Cheng, Newman and Bradley, Joshua and Chao, Alex and Mody, Apurva and Truitt, Steven and Metropolitansky, Dasha and Ness, Robert Osazuwa and Larson, Jonathan},
  journal={arXiv preprint arXiv:2404.16130},
  year={2024}
}

@article{guo2024lightrag,
  title={Lightrag: Simple and fast retrieval-augmented generation},
  author={Guo, Zirui and Xia, Lianghao and Yu, Yanhua and Huang, Chao},
  journal={Findings of the Association for Computational Linguistics: EMNLP 2025},
  year={2025}
}

@inproceedings{li2024graphreader,
  title={Graphreader: Building graph-based agent to enhance long-context abilities of large language models},
  author={Li, Shilong and He, Yancheng and Guo, Hangyu and Bu, Xingyuan and Bai, Ge and Liu, Jie and Liu, Jiaheng and Qu, Xingwei and Li, Yangguang and Ouyang, Wanli and others},
  booktitle={Findings of the Association for Computational Linguistics: EMNLP 2024},
  pages={12758--12786},
  year={2024}
}

@inproceedings{sarthi2024raptor,
  title={Raptor: Recursive abstractive processing for tree-organized retrieval},
  author={Sarthi, Parth and Abdullah, Salman and Tuli, Aditi and Khanna, Shubh and Goldie, Anna and Manning, Christopher},
  booktitle={International Conference on Learning Representations},
  volume={2024},
  pages={32628--32649},
  year={2024}
}

@article{gutierrez2024hipporag,
  title={Hipporag: Neurobiologically inspired long-term memory for large language models},
  author={Guti{\'e}rrez, Bernal J and Shu, Yiheng and Gu, Yu and Yasunaga, Michihiro and Su, Yu},
  journal={Advances in neural information processing systems},
  volume={37},
  pages={59532--59569},
  year={2024}
}

@inproceedings{anokhin2024arigraph,
  title={AriGraph: learning knowledge graph world models with episodic memory for LLM agents},
  author={Anokhin, Petr and Semenov, Nikita and Sorokin, Artyom and Evseev, Dmitry and Kravchenko, Andrey and Burtsev, Mikhail and Burnaev, Evgeny},
  booktitle={Proceedings of the Thirty-Fourth International Joint Conference on Artificial Intelligence},
  pages={12--20},
  year={2025}
}

@inproceedings{besta2024graph,
  title={Graph of thoughts: Solving elaborate problems with large language models},
  author={Besta, Maciej and Blach, Nils and Kubicek, Ales and Gerstenberger, Robert and Podstawski, Michal and Gianinazzi, Lukas and Gajda, Joanna and Lehmann, Tomasz and Niewiadomski, Hubert and Nyczyk, Piotr and others},
  booktitle={Proceedings of the AAAI conference on artificial intelligence},
  number={16},
  pages={17682--17690},
  year={2024}
}

@article{yao2023tree,
  title={Tree of thoughts: Deliberate problem solving with large language models},
  author={Yao, Shunyu and Yu, Dian and Zhao, Jeffrey and Shafran, Izhak and Griffiths, Tom and Cao, Yuan and Narasimhan, Karthik},
  journal={Advances in neural information processing systems},
  volume={36},
  pages={11809--11822},
  year={2023}
}

@article{cai2023temporal,
  title={Temporal knowledge graph completion: A survey},
  author={Cai, Borui and Xiang, Yong and Gao, Longxiang and Zhang, He and Li, Yunfeng and Li, Jianxin},
  journal={arXiv preprint arXiv:2201.08236},
  year={2022}
}

@inproceedings{trivedi2017knowevolve,
  title={Know-evolve: Deep temporal reasoning for dynamic knowledge graphs},
  author={Trivedi, Rakshit and Dai, Hanjun and Wang, Yichen and Song, Le},
  booktitle={international conference on machine learning},
  pages={3462--3471},
  year={2017}
}

@article{allen1983maintaining,
  title={Maintaining knowledge about temporal intervals},
  author={Allen, James F},
  journal={Communications of the ACM},
  volume={26},
  number={11},
  pages={832--843},
  year={1983},
  publisher={ACM New York, NY, USA}
}

@inproceedings{qin2021timedial,
  title={TIMEDIAL: Temporal commonsense reasoning in dialog},
  author={Qin, Lianhui and Gupta, Aditya and Upadhyay, Shyam and He, Luheng and Choi, Yejin and Faruqui, Manaal},
  booktitle={Proceedings of the 59th Annual Meeting of the Association for Computational Linguistics and the 11th International Joint Conference on Natural Language Processing (Volume 1: Long Papers)},
  pages={7066--7076},
  year={2021}
}

@inproceedings{xu2024knowledgeconflicts,
  title={Knowledge conflicts for llms: A survey},
  author={Xu, Rongwu and Qi, Zehan and Guo, Zhijiang and Wang, Cunxiang and Wang, Hongru and Zhang, Yue and Xu, Wei},
  booktitle={Proceedings of the 2024 Conference on Empirical Methods in Natural Language Processing},
  pages={8541--8565},
  year={2024}
}

@article{wang2023resolving,
  title={Resolving knowledge conflicts in large language models},
  author={Wang, Yike and Feng, Shangbin and Wang, Heng and Shi, Weijia and Balachandran, Vidhisha and He, Tianxing and Tsvetkov, Yulia},
  journal={arXiv preprint arXiv:2310.00935},
  year={2023}
}

@inproceedings{pham2024whoswho,
  title={Who’s who: Large language models meet knowledge conflicts in practice},
  author={Pham, Quang Hieu and Ngo, Hoang and Tuan, Luu Anh and Nguyen, Dat Quoc},
  booktitle={Findings of the Association for Computational Linguistics: EMNLP 2024},
  pages={10142--10151},
  year={2024}
}

@misc{langmem2026software,
  author       = {{The LangChain Team}},
  title        = {{LangMem SDK} for Agent Long-Term Memory},
  year         = {2025},
  month        = feb,
  howpublished = {\url{https://www.langchain.com/blog/langmem-sdk-launch}},
}

@misc{letta2026software,
  author       = {{Letta AI}},
  title        = {Letta: A Platform for Stateful Agents with Persistent Memory},
  year         = {2026},
  howpublished = {Software repository},
  url          = {https://github.com/letta-ai/letta},

}

@article{li2025memos,
  title={Memos: A memory os for ai system},
  author={Li, Zhiyu and Xi, Chenyang and Li, Chunyu and Chen, Ding and Chen, Boyu and Song, Shichao and Niu, Simin and Wang, Hanyu and Yang, Jiawei and Tang, Chen and others},
  journal={arXiv preprint arXiv:2507.03724},
  year={2025}
}

@inproceedings{yao2023react,
  title={ReAct: Synergizing Reasoning and Acting in Language Models},
  author={Yao, Shunyu and Zhao, Jeffrey and Yu, Dian and Du, Nan and Shafran, Izhak and Narasimhan, Karthik and Cao, Yuan},
  booktitle={International Conference on Learning Representations},
  year={2023}
}

@inproceedings{hmem2026,
  title={Hierarchical Memory for High-Efficiency Long-Term Reasoning in LLM Agents}, 
  author={Haoran Sun and Shaoning Zeng},
  booktitle={arXiv preprint arXiv:2507.22925},
  year={2025}
}

@inproceedings{hierarchical_graph_memory2026,
  title={Gam: Hierarchical graph-based agentic memory for llm agents},
  author={Wu, Zhaofen and Zhang, Hanrong and Lin, Fulin and Xu, Wujiang and Xu, Xinran and Chen, Yankai and Zou, Henry Peng and Chen, Shaowen and Zhang, Weizhi and Liu, Xue and others},
  booktitle={Proceedings of the 64th Annual Meeting of the Association for Computational Linguistics},
  pages={34647--34664},
  year={2026}
}

@article{gmemory2025,
  title={G-memory: Tracing hierarchical memory for multi-agent systems},
  author={Zhang, Guibin and Fu, Muxin and Wang, Kun and Wan, Frank and Yu, Miao and Yan, Shuicheng},
  journal={Advances in Neural Information Processing Systems},
  volume={38},
  pages={12988--13018},
  year={2025}
}

@inproceedings{kang2025memoryos,
  title={Memory os of ai agent},
  author={Kang, Jiazheng and Ji, Mingming and Zhao, Zhe and Bai, Ting},
  booktitle={Proceedings of the 2025 Conference on Empirical Methods in Natural Language Processing},
  pages={25972--25981},
  year={2025}
}

@inproceedings{meng2023memit,
  title={Mass-editing memory in a transformer},
  author={Meng, Kevin and Sharma, Arnab Sen and Andonian, Alex J and Belinkov, Yonatan and Bau, David},
  booktitle={The eleventh international conference on learning representations},
  year={2023}
}

@inproceedings{meng2022rome,
  title={Locating and editing factual associations in gpt},
  author={Meng, Kevin and Bau, David and Andonian, Alex J and Belinkov, Yonatan},
  booktitle={Advances in neural information processing systems},
  year={2022}
}

@inproceedings{chen2024lifelong,
  title={Lifelong knowledge editing for llms with retrieval-augmented continuous prompt learning},
  author={Chen, Qizhou and Zhang, Taolin and He, Xiaofeng and Li, Dongyang and Wang, Chengyu and Huang, Longtao and others},
  booktitle={Proceedings of the 2024 Conference on Empirical Methods in Natural Language Processing},
  pages={13565--13580},
  year={2024}
}

@inproceedings{mitchell2022serac,
  title={Memory-based model editing at scale},
  author={Mitchell, Eric and Lin, Charles and Bosselut, Antoine and Manning, Christopher D and Finn, Chelsea},
  booktitle={International Conference on Machine Learning},
  pages={15817--15831},
  year={2022}
}


\end{document}